\documentclass[letterpaper, 10 pt, conference]{ieeeconf}  

\IEEEoverridecommandlockouts                              

\let\labelindent\relax
\usepackage{enumitem}
\usepackage{hyperref}
\usepackage{xcolor}
\usepackage{color}
\usepackage{graphics} 
\usepackage{epsfig} 
\usepackage{mathptmx} 
\usepackage{times} 
\usepackage{amsmath} 
\usepackage{amssymb}  
\usepackage{subfig}
\usepackage{multirow}
\usepackage{algorithm2e}
\usepackage{romannum}
\usepackage{kpfonts}
\usepackage{todonotes} 
\usepackage{float}
\usepackage{booktabs}
\usepackage[font=small,labelfont=bf]{caption}
\usepackage[normalem]{ulem}
\usepackage{url}
\usepackage{threeparttable}

\DeclareUrlCommand\ULurl{%
  \renewcommand\UrlLeft{\uline\bgroup}%
  \renewcommand\UrlRight{\egroup}}

\renewcommand{\baselinestretch}{1}
\newcommand{\eg}{\emph{e.g. }}
\newcommand{\etc}{\emph{etc}}
\newcommand{\ie}{\emph{i.e. }}

\makeatletter
\def\BState{\State\hskip-\ALG@thistlm}
\makeatother

\makeatletter
\renewcommand{\paragraph}{%
  \@startsection{paragraph}{4}%
  {\z@}{0ex \@plus 0ex \@minus 0ex}{-1em}%
  {\hskip\parindent\normalfont\normalsize\bfseries}%
}
\makeatother

\title{\LARGE \bf 
Optimization Based Motion Planning for Multi-Limbed Vertical Climbing Robots
}

\author{Xuan Lin$^{1}$ Jingwen Zhang$^{1}$ Junjie Shen$^{1}$ Gabriel Fernandez$^{1}$ Dennis W Hong$^{1}$
\thanks{$^{1}$Robotics and Mechanisms Laboratory (RoMeLa), Department of Mechanical and Aerospace Engineering, University of California Los Angeles, CA 90095.
{\tt\small maynight@ucla.edu, zhjwzhang@g.ucla.edu, junjieshen@ucla.edu, gabriel808@g.ucla.edu, dennishong@ucla.edu}}%
}

\begin{document}

\maketitle
\thispagestyle{empty}
\pagestyle{empty}

\begin{abstract}
Motion planning trajectories for a multi-limbed robot to climb up walls requires a unique combination of constraints on torque, contact force, and posture. This paper focuses on motion planning for one particular setup wherein a six-legged robot braces itself between two vertical walls and climbs vertically with end effectors that only use friction. Instead of motion planning with a single nonlinear programming (NLP) solver, we decoupled the problem into two parts with distinct physical meaning: torso postures and contact forces. The first part can be formulated as either a mixed-integer convex programming (MICP) or NLP problem, while the second part is formulated as a series of standard convex optimization problems. Variants of the two wall climbing problem \eg, obstacle avoidance, uneven surfaces, and angled walls, help verify the proposed method in simulation and experimentation.
\end{abstract}


\section{Introduction}
Vertical wall climbing robots are applicable in many situations, such as surveillance, search and rescue, and building maintenance. Since wheeled vehicles can move fast on flat surfaces, wheeled robots have been experimented with for wall climbing \cite{beardsley2015vertigo} \cite{murphy2007waalbot}. Like many wheeled robots, non-legged wall-climbing robots are impeded by uneven surfaces, limiting their capabilities. Many animals found in nature demonstrate fast and agile climbing with their limbs. Legged animals can climb up highly unstructured environments as well as traverse on ground. They are also able to jump onto and grab structures using only their hands, demonstrating highly mobile motions that non-legged robots cannot even attempt. Much of the research on climbing robots started by mimicking animals \cite{kim2008smooth} \cite{spenko2008biologically}, and then gradually the systems became more complex. The most recent advancement is \cite{parness2017lemur} which presented a 35 kg robot with 4, 7-degree-of-freedom limbs, climbing on smooth surfaces with a gecko type gripper and rough surfaces with micro-spine gripper.

When generating climbing motions, many researchers in the field resort to templates \cite{lynch2012bioinspired} \cite{clark2007design}. Templates are used to study the dynamics of climbing. Its implementation is currently limited to light, low degree-of-freedom robots. The climbing motion for more complex, high degree-of-freedom robots is still quasi-static \cite{parness2017lemur}. When climbing an environment not seen before, the robot needs to carefully plan its steps and torque based on the environment to find a trajectory to reach its objective. Thus, wall climbing becomes a motion planning problem. This line of work was started by \cite{bretl2006motion}, which presented an algorithm based on a more classical graph search method. In this paper, we present another approach which utilizes optimization methods for multi-limbed climbing robots to plan trajectories when climbing. Optimization based methods, such as mixed-integer convex programming (MICP) and nonlinear programming (NLP), have been implemented in many situations to plan motions for walking robots \cite{kuindersma2016optimization} \cite{winkler2018gait} \cite{ahn2018stable} \cite{aceituno2018simultaneous}. This paper extends these methods to wall-climbing applications.

\begin{figure}[!t]
		\centering
		\includegraphics[scale=0.71]{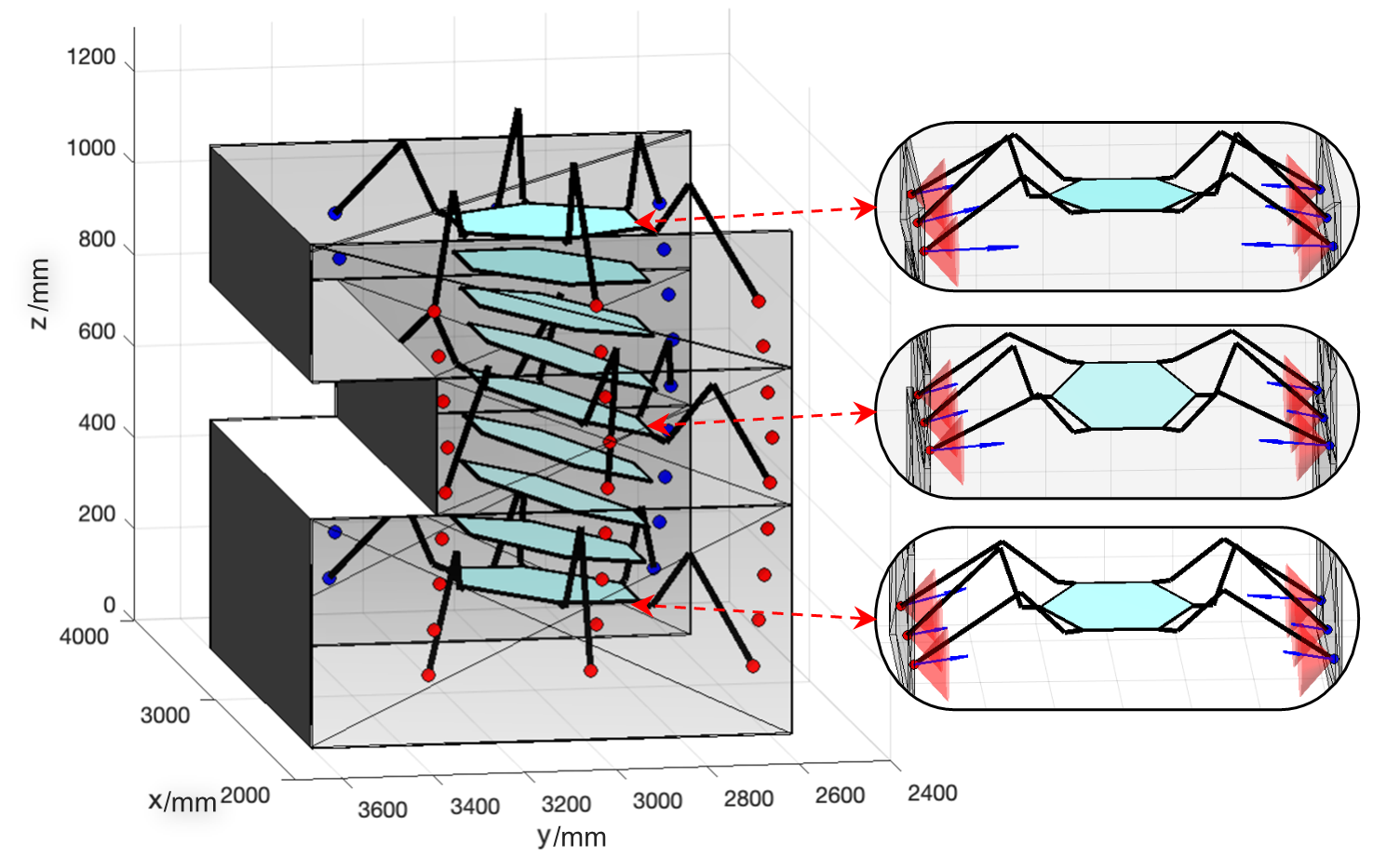}
		\caption {A series of trajectories generated by the MICP motion planner for the six-legged robot to climb up between two walls while avoiding an obstacle. Red and blue dots show the planned toe positions, and the hexagons show the body orientation. Three postures are detailed on the right with planned contact forces (blue arrows) along with the nominal friction cones (red).}
		\label{Fig: bracing plot}
\end{figure}


One difference between planning for walking versus for climbing is that for walking, the planner focuses on generating a series of poses that satisfy various constraints. However, for climbing, posture constraints alone is insufficient. Each end effector needs to preload before making contact with a surface in order to generate enough adhesive forces. For example, geckos needs to preload their foot with shear forces to generate normal forces \cite{autumn2006dynamics}. Thus, for each different posture the robot needs to reason: how much preload needs to be exerted? Is the adhesive force sufficient to hold the robots weight without slipping? Are the actuators strong enough to keep the pose on the wall and to preload the end effectors? To answer these questions, additional constraints need to be reasoned, leading to a different algorithm from walking.


The algorithm presented in this paper is experimentally verified by enabling a position controlled hexapod robot to climb between two walls with frictional contacts, \ie \emph{two-wall-climbing} problem. This problem was first introduced by our previous paper \cite{xuan2018multi}. Although the robot doesn't use any adhesive-type grippers, it ``squeezes" itself between two walls, like loading a spring, to generate a large normal force. Therefore, the ``preloading" described above corresponds to squeezing the body. There are a few key characteristics for this sort of climbing. First, the problem is quasi-static but statically indeterminate. Thus, a different model based on stiffness matrices is required to solve the contact force, as presented in \cite{xuan2018multi}. Second, since this robot utilizes no more than pure friction to climb, significant amounts of pushing force against the wall is required by the limbs. This constantly drives the actuators close to its maximum output torque. Our algorithm makes a trade-off between the pushing force and motor torque. If a limb pushes too hard into the wall, it may over-torque itself, while if it doesn't push hard enough, the friction may not be sufficient to hold the robot's weight. This inspired a safety factor based design method, as presented in the next section. The results of this paper may be directly transcribed to certain kinds of human-like climbing, \eg, tree-climbing, climbing down a well, \etc.



This paper makes the following contributions:
\begin{enumerate}[leftmargin=*]
\item A two-step optimization based motion planning algorithm for multi-limbed vertical climbing robots:
\begin{enumerate}[leftmargin=*]
\item An MICP formulation for climbing posture planning.
\item A convex optimization formulation for contact force planning when climbing with the proposed safety factor.
\end{enumerate}
\item Demonstration of the algorithm on hardware, \ie, a hexapod robot that climbs between two irregular walls with frictional contacts.


\end{enumerate}

\section{Problem Formulation}
This section describes the robotic platform used for climbing, the model of the platform, and a complete formulation of the optimization problem.
\subsection{Robotic Platform}

\begin{table}[b!]
\vspace{-9pt}
\caption{Robot Configuration}
\vspace{-12pt}
\label{spec}
\begin{center}
\resizebox{0.7\linewidth}{!}{
\begin{tabular}{c|c}
\hline
\hline
Parameter & Value \\
\hline
Degree of Freedom for Each Limb & 3 \\
Limb Coxa Length & 57 [mm] \\
Limb Femur Length & 195 [mm] \\
Limb Tibia Length & 375 [mm] \\
Weight & 10.3 [kg]\\
Max Torque & 26 [Nm] \\
\hline
\hline
\end{tabular}}
\end{center}
\end{table}
The robotic platform used in this paper is a hexapod robot whose each limb has 3 degrees of freedom and consists of a coxa, an upper femur, and a lower tibia assembly. Dynamixel MX-106 motors have been used in pairs for actuation. The robot carries its own battery, computer, and IMU. End effectors are covered by anti-slip tape to enhance friction. The parameters of the robot are summarized in Table~\ref{spec}. For more details of the robot design see \cite{xuan2018multi}.

\subsection{Robot Model}
\begin{figure}[t]
		\centering
		\includegraphics[scale=0.3]{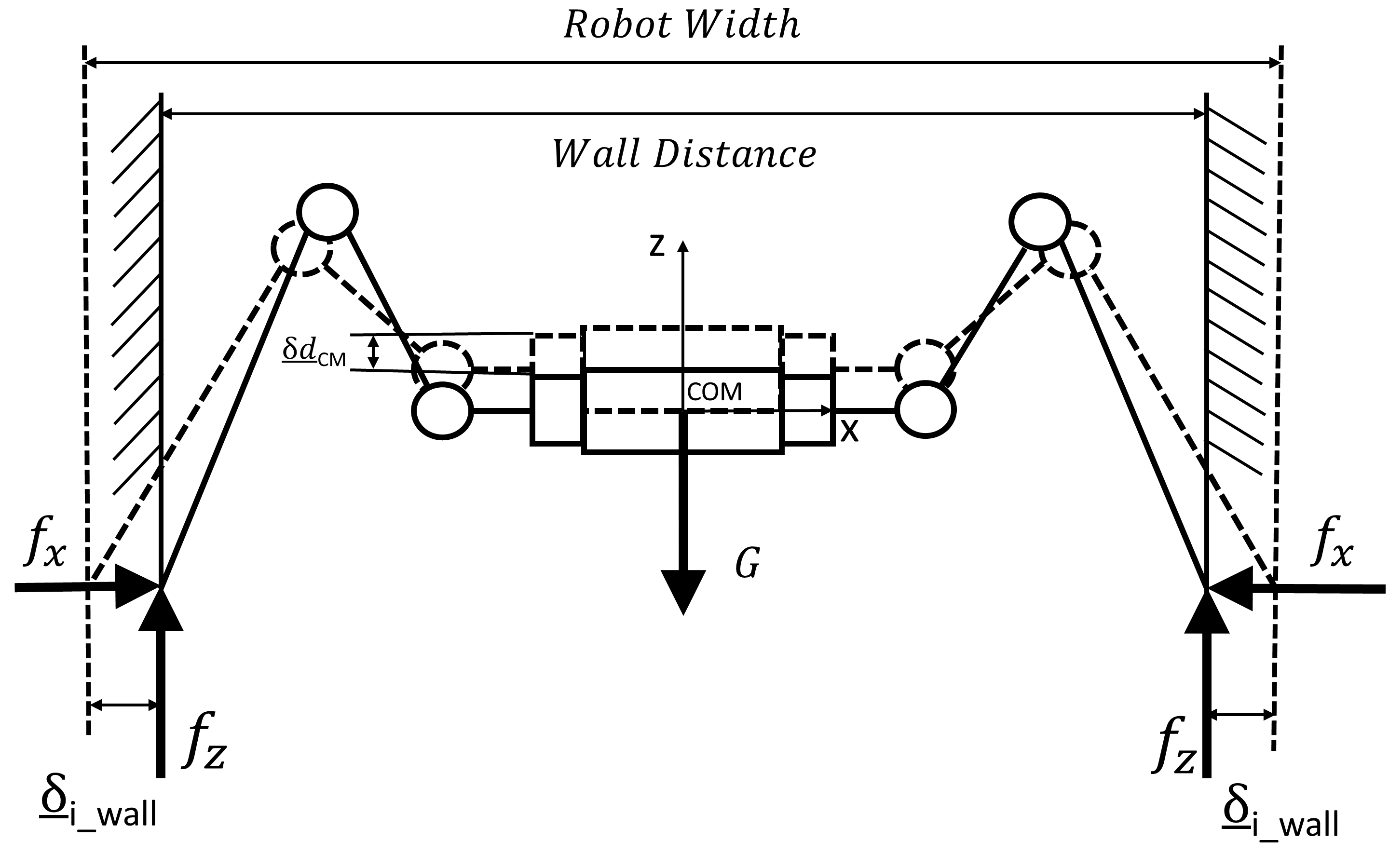}
		\caption {Deflection of hexapod bracing between walls}
		\label{Fig: deflection}
\end{figure}

One difficulty to model multi-limbed vertical climbing robots with position controlled joints is the reaction forces are \emph{statically indeterminate} \cite{kumar1990force} \cite{hong2006visualization}, \ie they cannot be completely determined by the static equilibrium equations when the robot makes more than 3 contact points on the environment. To calculate the contact force completely, one needs to consider the deformation of the robotic system. This is especially true for \emph{two-wall-climbing} robots, since the normal reaction forces ($f_{x}$ in Fig. \ref{Fig: deflection}) are mainly statically indeterminate. This problem has been mostly ignored in previous literature but studied in our last paper \cite{xuan2018multi}, using Virtual Joint Method (VJM) to model limb compliance and summarize into a \emph{whole body stiffness matrix}. The results are presented below. For detailed derivations see \cite{xuan2018multi}.

Using the VJM method, the stiffness matrix for a L-degree-of-freedom limb is given by:

\begin{equation}
\textbf{K} = (\textbf{J}\textbf{k}^{-1}\textbf{J}^{T})^{-1}
\label{Eq:stiffness matrix - result}
\end{equation}
where:
\begin{equation}
\textbf{k} = \textrm{diag}(k_{i}), \;\; i=1, \dots,L
\end{equation}
is a diagonal matrix composed of the spring coefficients of the P-controlled servos. $\textbf{J}$ is a $3 \times L$ Jacobian matrix.

When the robot with $N$ limbs is bracing between two walls, its center of mass has a small deflection, \emph{sagdown}, denoted by $\underline{\delta}_{COM}=[\underline{\delta d}_{COM}, \underline{\delta\theta}_{COM}]^{T}$. The wall squeezes the robot, imposing a deflection denoted by $\underline{\delta}_{i\_ wall}$. The torso's center of mass deflection can be related by the wall imposed deflections and external load $[\underline{F}_{tot}, \underline{M}_{tot}]^{T}$ (in this case only gravity) through:

\begin{equation}
\textbf{A}\underline{\delta}_{COM}
=
\begin{bmatrix}
F_{tot} \\
M_{tot} \\
\end{bmatrix}
+
\sum^{N}_{i=1}
\begin{bmatrix}
\textbf{K}_{i}      \\
\textbf{P}_{i}\textbf{K}_{i} \\
\end{bmatrix}
\underline{\delta}_{i\_ wall} \\
\label{Eq:whole body stiffness model}
\end{equation}
where:
\begin{equation}
\textbf{A}=\sum_{i=1}^{N}
\begin{bmatrix}
\textbf{K}_{i}                &  \textbf{K}_{i}\textbf{P}_{i}^{T}\\
\textbf{P}_{i}\textbf{K}_{i}  &  \textbf{P}_{i}\textbf{K}_{i}\textbf{P}_{i}^{T}\\
\end{bmatrix}
\label{Eq:whole body stiffness matrix}
\end{equation}

\begin{equation}
\textbf{P}_{i} =
\begin{bmatrix}
0      &  -z_{i}   &   y_{i}   \\
z_{i}   &     0     &   -x_{i}  \\
-y_{i}  &    x_{i}  &     0     \\
\end{bmatrix}
\label{Eq:Cross Product Matrix}
\end{equation}
is the anti-symmetric matrix from each toe position.

And the reacting force on each toe is given by:

\begin{equation}
\underline{f}_{i}=\textbf{K}_{i}(\underline{\delta}_{i\_ wall}-[\textbf{I}\;\; 
\textbf{P}_{i}^{T}]\underline{\delta }_{COM}), \;\; i=1,\dots,N
\label{Eq:End effector reaction force}
\end{equation}

\subsection{Safety Factor for Climbing}
\label{Sec:safety_factor}

Another difference in wall climbing from walking is that wall climbing is a high-risk task. Falling down from a climb is likely to not only damage the robot and its environment but also injure people. When in a non-controlled environment, several uncertainties may cause climbing to fail unexpectedly. For instance, the friction coefficient can never be measured precisely, or there may be unexpected external load \eg, wind. In our analysis wall-climbing tasks are typically static postures since the process happens slowly. However, there still exist velocities which may cause the end effector to disengage or over-torque. For this reason the authors propose the notion of safety factors for wall-climbing motions. In order to motion plan, it not only needs to satisfy the nominal constraint but also needs to satisfy the safety factor constraint.


Similar to finite element analysis, the safety factor is generated by analyzing each posture of the motion and calculating the ratio of the current index over the critical failure index. In this paper, we investigate a robot climbing between two walls with frictional contact, and the two failure modes are insufficient friction (slip) and motor over-torque. Therefore, there are two safety factors to consider. Imagine being able to gradually reduce $\mu$ from the nominal value to the critical value $\mu_{c}$, when the robot is about to slip. This provides us with a notion of the safety factor with respect to the coefficient of friction, $S_{\mu}$, defined in equation (7). Similarly, if we imagine lowering the motor torque limit $\tau_{max}$ from its nominal value to a critical value $\tau_{c}$, which is right before the motor over-torques. We can define another safety factor with respect to the motor's max torque, $S_{\tau}$, defined by equation (8).



\begin{align}
S_{\mu}&=\mu/\mu_{c} \\
S_{\tau}&=\tau_{max}/\tau_{c}
\label{Eqn:friction_required}
\end{align}

These notions are first introduced in our previous paper \cite{xuan2018multi}, where it can be retrieved graphically from feasibility region analysis. In this paper, we formulate a convex optimization problem to plan for the amount of pushing forces that satisfy the safety factor constraints for each planned robot pose.

\subsection{Complete Formulation of the Planning Problem}
\begin{figure*}[t]
		\centering
		\includegraphics[scale=0.62]{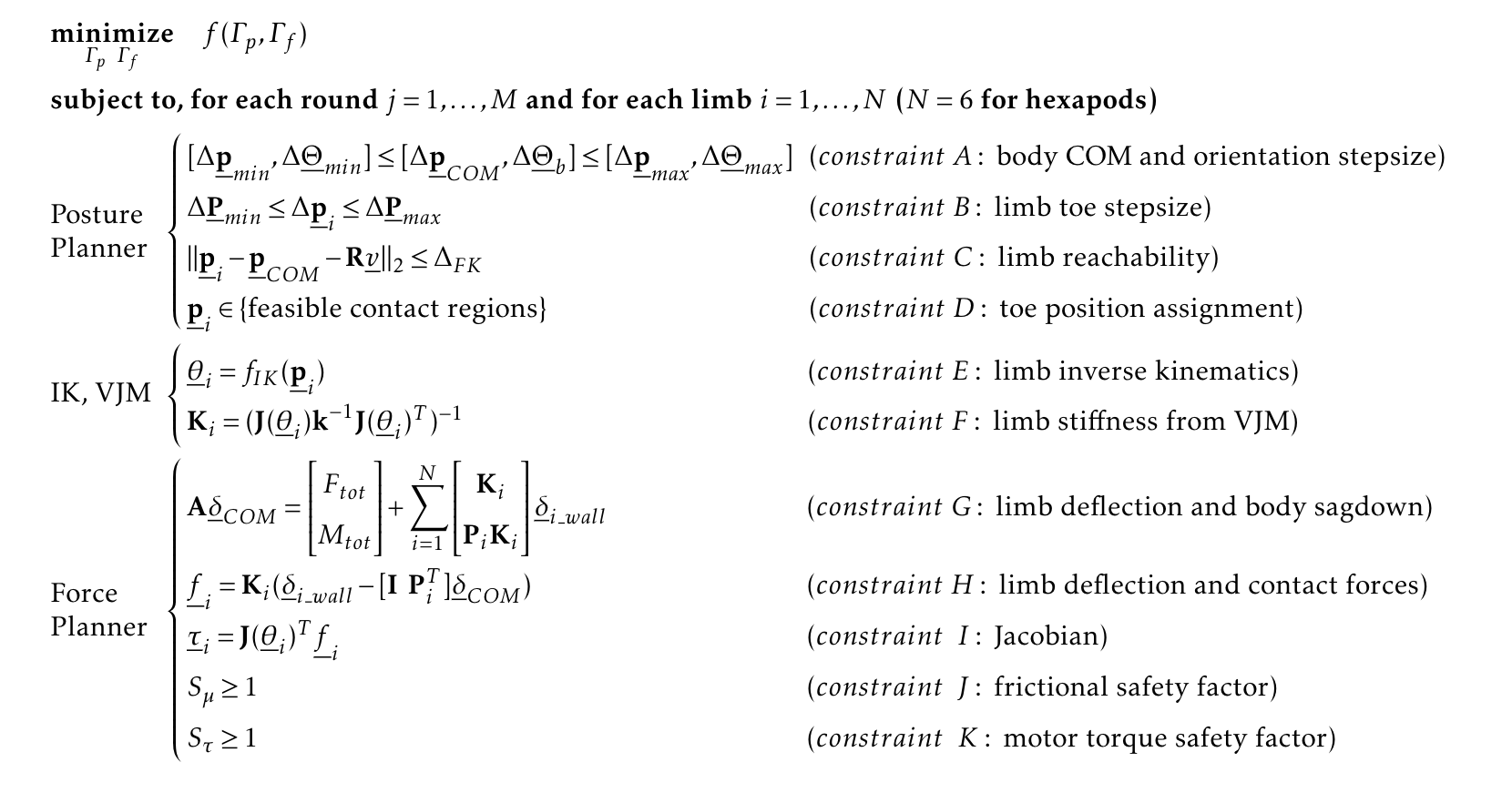}
		\caption {Complete optimization formulation of the motion planning problem}
		\label{Fig:complete_formulation}
\end{figure*}
In Fig. \ref{Fig:complete_formulation}, we present here the complete mathematical formulation of motion planning problem for $M$-rounds climbing between walls with friction, where part of the decision variables, $\mathit{\Gamma}_{p}$, are

\begin{equation}
    \mathit{\Gamma}_{p} = \{  \textbf{\underline{p}}_{i}[j], \ \textbf{\underline{p}}_{COM}[j], \ \underline{\Theta}_{b}[j] \ | \ i = 1,\ldots,\textit{N}, \ j = 1,\ldots,\textit{M}  \}
\end{equation}\\
and the other part of decision variables, $\mathit{\Gamma}_{f}$, are

\begin{equation}
\begin{aligned}
    \mathit{\Gamma}_{f} = \{ \ & \underline{\delta }_{COM}[j], \ \underline{f}_{i}[j], \ \underline{\delta}_{i\_ wall}[j], \ \textbf{K}_{i}[j] \\
    & \ \ | \ i = 1,\dots,\textit{N}, \ j = 1,\ldots,\textit{M} \ \}
\end{aligned}
\end{equation}
for each round $j$, it plans body's center of mass (COM) position $\textbf{\underline{p}}_{COM}[j]$, body orientation ${\Theta}_{b}[j]$, body deflection $\underline{\delta}_{COM}[j]$ and the $i^{th}$ limb's toe positions $\textbf{\underline{p}}_{i}[j]$, the limb stiffness matrices $\textbf{K}_{i}[j]$, the contact forces $\underline{f}_{i}[j]$, and limb deflections $\underline{\delta}_{i\_wall}[j]$, where $i$ is the limb index.

\textit{Constraint} A and B limit the range of travel between rounds. In \textit{constraint} C, we approximate the limb workspace by a ball. $\underline{v}$ is the vector from robot body's COM to the first joint of the limb. \textit{Constraint} D ensures the toe lies on a feasible contact region on the wall. In this paper, we assume perfect knowledge of the wall geometry, \ie its mathematical expression is available to the planner. \textit{Constraint} E represents inverse kinematics. \textit{Constraint} F is equation (\ref{Eq:stiffness matrix - result}), the limb stiffness matrix based on VJM. \textit{Constraint} G is equation (\ref{Eq:whole body stiffness model}) (\ref{Eq:whole body stiffness matrix}) (\ref{Eq:Cross Product Matrix}), the whole body stiffness model. \textit{Constraint} H is equation (\ref{Eq:End effector reaction force}) which relates limb contact force with its deflection.
\textit{Constraint} H, I, J, and K ensure the safety factor constraint in Section \ref{Sec:safety_factor} is satisfied.

Given the results from \cite{winkler2018gait}, the complete problem may be solvable with a single NLP solver. Instead of doing that, we chose to separate the problem into two parts that solve an MICP/NLP problem first and then solve a series of standard convex optimization problems, as demonstrated in the next section.

\section{Motion Planning Algorithm}

In this section, we describe how we solve the optimization problem in the last section. Several papers \eg, \cite{winkler2018gait}, have demonstrated the power of NLP solvers being able to solve various nonlinear motion planning problems. However, NLP solvers can easily get trapped by local minima if the problem is complicated. We noticed that in the optimization problem shown in Fig. \ref{Fig:complete_formulation}, \textit{constraint} F can be numerically hard for optimization solvers, especially due to potential singularity issues. Therefore, we chose to naturally decouple the problem into two sections at \textit{constraint} F, as indicated on the left of Fig. \ref{Fig:complete_formulation}. The first section is composed of \textit{constraint} A, B, C, D. This is similar to a standard walking motion planning problem and can be solved by an MICP or NLP solver given the vast existing literature. The second section including \textit{constraint} G, H, I, J, K. Given all toe positions and body posture, this part is a standard force distribution problem \cite{kumar1990force}, and can be formulated into a standard convex optimization problem that can be easily solved. In this setup, the \textit{constraint} F along with \textit{constraint} E are evaluated algebraically after solving the first section of the problem and are not fed into any optimization solver. By doing so, we sacrifice some optimality, which can be seen in the following form:

\begin{equation*}
\begin{aligned}
& \underset{\mathit{\Gamma}_{p} \,\ \mathit{\Gamma}_{f}} {\text{minimize}} && f(\mathit{\Gamma}_{p},\mathit{\Gamma}_{f})\\
& \text{subject to} && h_{1}(\mathit{\Gamma}_{p}) \le 0 \\
&&& h_{2}(\mathit{\Gamma}_{p}, \mathit{\Gamma}_{f}) \le 0 \\
\end{aligned}
\end{equation*}

The constraint $h_{1}\le0$ denotes the part that is to be solved in the first part of the optimization problem while $h_{2}\le0$ in the second part of the problem. Compared to an NLP solver that takes care of both sets of constraints simultaneously, in our 2-step setup constraint $h_{1}\le0$ is solved independent of $h_{2}\le0$. This means an optimal solution to $h_{1}\le0$ may render $h_{2}\le0$ non-optimal. However, we choose to solve this problem in such way since it has several advantages: 

\begin{enumerate}
    \item \textbf{Interpretability} The two problems have clear and distinct physical interpretations. The first part focuses on solving a series of postures, while the second part optimizes for how much force the robot needs to exert on the wall. If at one round the solver fails, it is clear why the planner fails, and part of the feasible solutions may still be used. Whereas for an NLP solver, if it returns \emph{infeasible}, it tends to return results that can't be utilized and with no interpretable information.
    
    \item \textbf{Adaptability} The first part of the problem is identical to the motion planning problem for legged walking. Thus, it easily connects to the vast literature of walking robot motion planning. Only the second part depends on end effectors for climbing, which can be easily reformulated if the end-effector is swapped.
    
    \item \textbf{Speed} Decoupling the problem into two parts turns a bulk part of the problem into convex optimization problems, which can be solved efficiently. This can be justified by Table \ref{Tab:MICPsolution_speed}. 
    
\end{enumerate}

In the next two sections, we introduce detailed formulations for each part of the problem.


\subsection{Optimization for Climbing Posture}
Given a goal configuration during wall climbing, a pre-defined number of postures $\textit{M}$ to reach it should be computed under the constraints of step size, kinematics, and toe contact points within feasible contact regions. To simplify the task of assigning toe contact points, we divided feasible contact regions into several pre-computed convex constraints represented by $\textbf{A}_{r}\textbf{p}_{i}\le\textbf{b}_{r}$ with perfect knowledge of structured wall geometry, where $r$ is the index of feasible contact region. The IRIS algorithm \cite{kuindersma2016optimization} used for typical walking robot motion planning problem, is also able to compute these regions with perception. The entire optimization problem for climbing posture is formulated as follows:

\begin{equation*}
\begin{aligned}
& \mathop {\text{minimize}}\limits_{\mathit{\Gamma}_{p}, \ \textbf{\textit{H}}} && {(\textbf{\underline{q}}[M] - {\textbf{\underline{q}}_g})^T}{\textbf{\underline{W}}_g}(\textbf{\underline{q}}[M] - {\textbf{\underline{q}}_g}) \ +  \\
&&& \sum\limits_{j = 1}^{M - 1} {({J_{COM}} + {J_{ROT}} + {J_S}}) \\
&\text{subject to} && \text{for $j = 1, \ldots, \textit{M}, $}\\
&&&\Delta\textbf{\underline{p}}_{min}
 \le \| \textbf{\underline{p}}_{COM}[j]-\textbf{\underline{p}}_{COM}[j-1] \|_{2} \le \Delta\textbf{\underline{p}}_{max} \\
&&&\Delta\textbf{\underline{P}}_{min} \le \| \textbf{\underline{p}}_{i}[j]-\textbf{\underline{p}}_{i}[j-1] \|_2 \le \Delta\textbf{\underline{P}}_{max} \\
&&&\Delta\underline{\Theta}_{min} \le \underline{\Theta}_{b}[j]-\underline{\Theta}_{b}[j-1] \le \Delta\underline{\Theta}_{max} \\
&&&\| \textbf{\underline{p}}_{i}[j]-\textbf{\underline{p}}_{COM}[j]-\textbf{R}\underline{v}\|_2 \le\Delta_{FK} \\
&&&H_{r,i}[j] \ \Rightarrow \ \textbf{A}_{r}\textbf{p}_{i}\le\textbf{b}_{r} \\
&&&\sum_{r=1}^{R}H_{r,i}[j]=1 \ \ H_{r,i}\in{0,1}
\end{aligned}
\end{equation*}

where $\Delta\textbf{\underline{p}}_{min}$, $\Delta\textbf{\underline{p}}_{max}$, $\Delta\textbf{\underline{P}}_{min}$, $\Delta\textbf{\underline{P}}_{max}$, $\Delta\underline{\Theta}_{min}$, $\Delta\underline{\Theta}_{max} \in {\mathbb{R}^3}$ are bounds for the toe, body's COM and orientation step sizes. $\Delta_{FK} \in \mathbb{R}$ is the radius of the limb workspace ball. For each round, $\textbf{\textit{H}} \in \{0, 1\}^{R \times 6}$ is taking on integer values to assign toes to feasible contact regions where $R$ is the number of feasible contact regions. The conditional constraint about $H_{r,i}$ is represented using a standard big-M formulation. 

In terms of cost function, the first term is introducing the distance of the last round from goal configuration where $\textbf{\underline{q}}[M] = [\textbf{\underline{p}}_1[M], \ \ldots, \ \textbf{\underline{p}}_6[M] ]$ while $\textbf{\underline{q}}_g$ is the goal configuration. The shifting amounts $J_{COM}$, $J_{ROT}$ and $J_{S}$ of body's COM, orientation and toe positions are added to avoid turning or climbing too far in one single step, as the second term of cost function:

\begin{equation*}
    \left \{ \begin{array}{lr}
    J_{COM} = \Delta\textbf{\underline{p}}_{COM}^T{\textbf{\underline{W}}_{COM}} \Delta\textbf{\underline{p}}_{COM} \\
     J_{S} = \sum\limits_{i = 1}^{6} {\Delta\textbf{\underline{p}}_{i}^T{\textbf{\underline{W}}_{s}}\Delta\textbf{\underline{p}}_{i}} \\
    J_{ROT} = \Delta\underline{\Theta}_{b}^T{\textbf{\underline{W}}_{ROT}}\Delta\underline{\Theta}_{b} 
    \end{array}
    \right.
\end{equation*}
where $\textbf{\underline{W}}_{COM}$, $\textbf{\underline{W}}_{s}$, $\textbf{\underline{W}}_{ROT}$ and previous $\textbf{\underline{W}}_{g}$ are weights used to tune the optimizer. Except rotation matrix $\textbf{R}$ computed from $\underline{\Theta}_{b}[j]$, which has a nonlinear constraint, other parts forms an MICP, since they are either linear or quadratic (convex). To address the nonlinearity, linear approximation of $\textbf{R}$ for $\underline{\Theta}_{b}[j] = [\alpha, \beta, \gamma]$ is used as follows:

\begin{equation}
\textbf{R}(\underline{\Theta}_{b}[j]) =
\begin{bmatrix}
1      &  -\gamma   &   \beta   \\
\gamma   &     1     &   -\alpha  \\
-\beta  &    \alpha  &     1     \\
\end{bmatrix}
\label{Eq:linear rotation}
\end{equation}


This approximation is valid for applications involving small rotations which is reasonable for wall-climbing applications that do not require large rotations. The error is under $3 \%$ while angles are smaller than $10^{\circ}$ using the Frobenius norm of a matrix to compare the similarity between linearly-approximated and exact rotation matrix.

When large changes in the body orientation are expected, linearization of the rotation matrix becomes invalid. In this case, NLP can be used. We formulated our NLP according to \cite{winkler2018gait} without considering the dynamic model. Moreover, NLP can easily handle non-convex terrains, \eg, round tubes, which extends the application of our work.

\subsection{Optimization for Pushing Force}
After the posture planner is finished, the inverse kinematics, \textit{constraint} E, and the stiffness matrix, \textit{constraint} F, may be evaluated directly. Since those constraints are complicated and may have numerical stability issues due to the inverse of the matrix, we avoid directly placing it in a gradient based solver. The second part of algorithm tackles the problem of how much force each limb needs to exert on the wall. We use a pre-defined gait to go from one planned posture to the next one. The robot lifts one leg and puts it on the wall, pushes the body upwards, then lifts another leg, and repeats. We pick 12 critical instants between two postures for the force planner to investigate: 6 instants after the robot lifts one leg and 6 instants after the robot pushes its body up. The planner is formulated into a series of standard convex optimization problem so that it can be solved efficiently.



In Section \ref{Sec:safety_factor} the notion of climbing motion safety factors, $S_{\mu}$ and $S_{\tau}$, are proposed. Since the two safety factors are inversely related: pushing harder against the wall will increase $S_{\mu}$ but decrease $S_{\tau}$ and vice versa. We would like to guarantee that the safety factors are above a value larger than 1 while having a weight to tune the pushing force. This can be formulated as:

\begin{equation*}
\begin{aligned}
& \underset{\tau_{i} \ \underline{f}_{i}} {\text{maximize}} && S_{\tau} + wS_{\mu} \\
& \text{subject to}&& |\tau_{i}| \le \tau_{max}/S_{\tau} \\
&&& \underline{n}_{i}^{T}\underline{f}_{i} \ge0 \\
&&& \|\underline{f}_{i}-(\underline{n}_{i}^{T}\underline{f}_{i})\underline{n}_{i}\|_{2}\le (\mu/S_{\mu}) (\underline{n}_{i}^{T}\underline{f}_{i}) \\
&&& S_{\mu}\ge1 \ , \ S_{\tau}\ge1
\end{aligned}
\end{equation*}

where $\underline{n}_{i}$ is the wall normal vector at toe $i$, and $w$ is the weight to trade-off between the two safety factors. If we define $S_{\tau\_inv}=1/S_{\tau}$, we can write the torque constraints into a linear form. Although the friction cone constraint itself is convex, adding in frictional safety factor $S_{\mu}$ as an optimization variable makes it non-convex. Thus we set a constant $S_{\mu}$, which shrinks the friction cone. We put normal reaction forces $\underline{n}_{i}^{T}\underline{f}_{i}$ into the objective function. By tuning the amount of normal reaction force, the effective friction cone constraint can be made looser or tighter. Stated formally:

\begin{equation*}
\begin{aligned}
&\underset{\underline{\delta }_{COM} \ \underline{f}_{i} \ \underline{\delta}_{i\_ wall} \ \tau_{i} \ S_{\tau\_inv}} {\text{minimize}} &&S_{\tau\_inv} - w\sum_{i=1}^{N}\underline{n}_{i}^{T}\underline{f}_{i}\\
& \text{~~~~~~subject to} &&\textbf{A}\underline{\delta}_{COM} =
\begin{bmatrix}
F_{tot} \\
M_{tot} \\
\end{bmatrix}
+
\sum^{N}_{i=1}
\begin{bmatrix}
\textbf{K}_{i}      \\
\textbf{P}_{i}\textbf{K}_{i} \\
\end{bmatrix}
\underline{\delta}_{i\_ wall} \\
&&& \underline{f}_{i}=\textbf{K}_{i}(\underline{\delta}_{i\_ wall}-[\textbf{I}\;\; 
\textbf{P}_{i}^{T}]\underline{\delta }_{COM})\\
&&& \underline{\tau}_{i} = \textbf{J}(\underline{\theta}_{i})^{T}\underline{f}_{i} \\
&&& 0\le S_{\tau\_inv}\le1\\
&&& |\tau_{i}| \le S_{\tau\_inv}\tau_{max} \\
&&& \underline{n}_{i}^{T}\underline{f}_{i} \ge0 \\
&&& \|\underline{f}_{i}-(\underline{n}_{i}^{T}\underline{f}_{i})\underline{n}_{i}\|_{2}\le \mu (\underline{n}_{i}^{T}\underline{f}_{i})/S_{\mu}
\end{aligned}
\end{equation*}
\medskip

Increasing $w$ makes the normal reaction force higher. This loosens the friction cone bound, since the required shear force ($f_z$ in Fig. \ref{Fig: deflection}) is static determinate (half of the gravity $G$). Decreasing $w$ increases $S_{\tau}$, while tightening the friction cone bound. This problem is convex and can be solved quickly with a global optimal guarantee. According to our trial and error hardware testing, a lower bound of $S_{\mu}=1.8$ provides sufficient safety against the coefficient of friction.

\section{Results}

We present here three scenarios that we investigated using our planner: climbing over  steps on the walls, climbing on the walls while avoiding obstacles, and climbing on non-parallel walls. All results are validated on actual hardware with properly tuned weights. In each experiment, the walls are covered by rubber pads and the robot toes are covered by anti-slip tape, which gives a frictional coefficient $\mu$ around 1. A body posture regulator based on IMU orientation feedback and PID control is utilized for the robot body to track the planned orientation. No other feedback is used. Due to the stable but slow one-leg gait, the climbing speeds in all cases are around 20 cm/min. All hardware demonstrations can be viewed in the accompanied video {\small({\color{blue}\underline{\url{https://www.youtube.com/watch?v=AXmrqnt3JIA&t=2s}}})}.

\subsection{Climbing over steps on the walls}
\begin{figure}[t]
		\centering
		\includegraphics[scale=0.33]{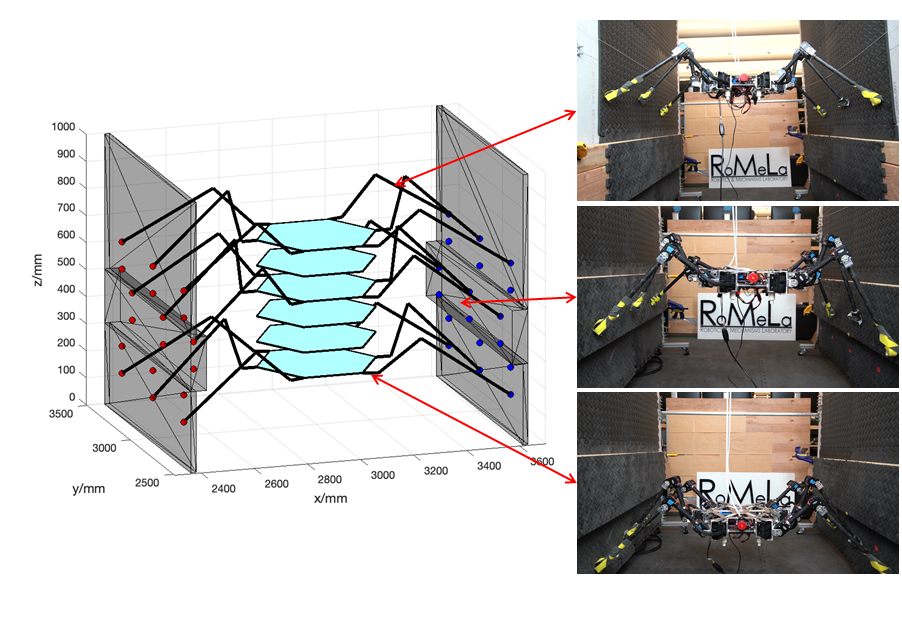}
		\caption {Visualization and hardware testing of planning results for climbing over steps on the walls.}
		\label{Fig:result_steps}
\end{figure}

In this scenario, we let the robot climb between two walls at a distance of 1230mm but with a 40mm thick by 200mm high step on both walls. The wall has multiple feasible contact surfaces, but the robot doesn't need to rotate for this task. Therefore, MICP is used for planning the robot posture with the body orientation kept flat. However, the robot does need to adapt to the tightening of the walls' distance between the walls. On the steps, the robot doesn't push as far out to prevent over-torque. Fig. \ref{Fig:result_steps} shows the planned series of postures, visualized in MATLAB, as well as associated hardware testing scenarios. 

\begin{figure}[t]
		\centering
		\makebox[0.35\textwidth][c]{
		\includegraphics[scale=0.37]{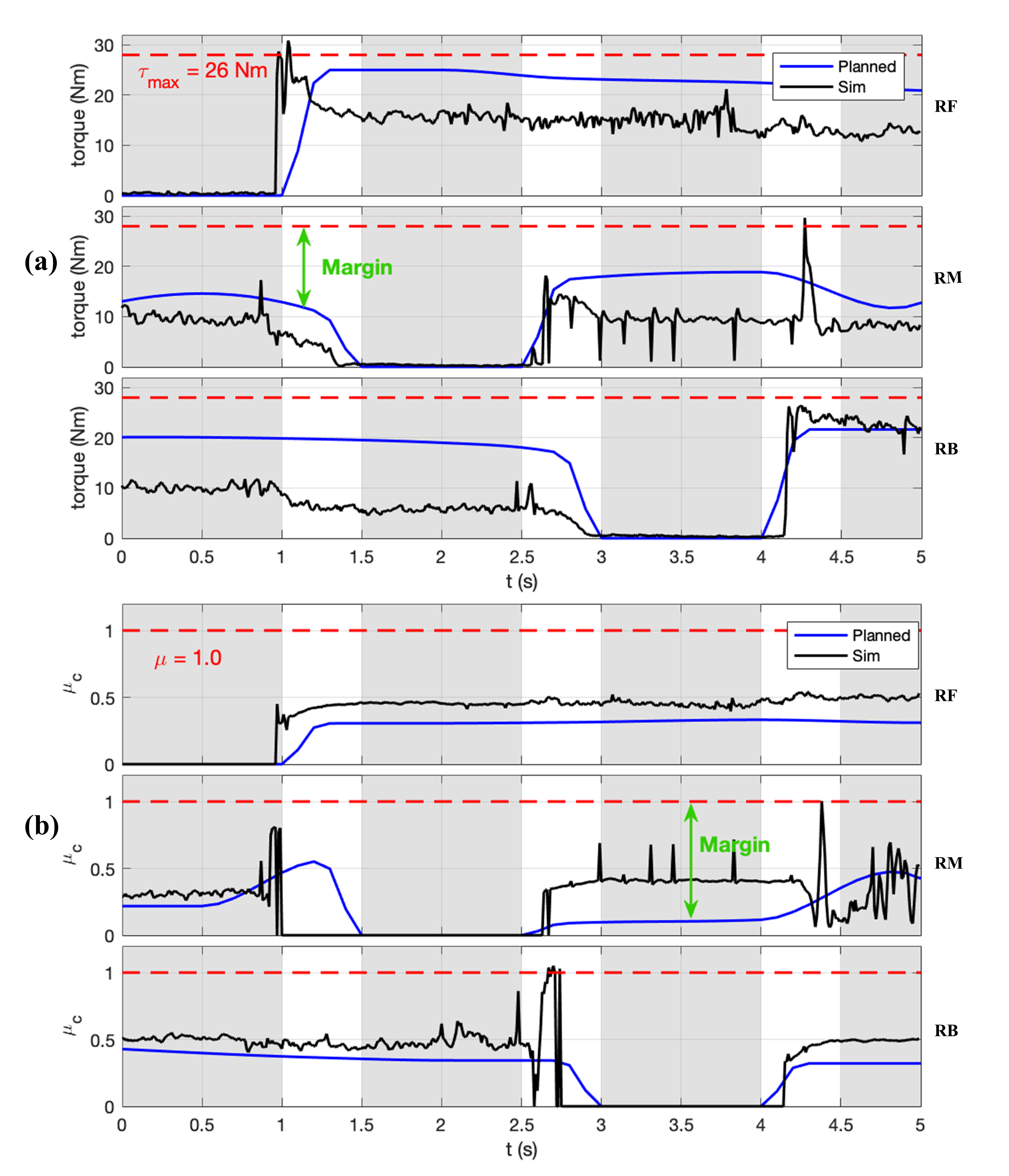}}
		\caption {Diagrams of the planned and V-rep (Bullet 2.83 engine) simulated results for the required motor torque $\tau_{c}$ (maximum of 3 motors, diagram (a)) and frictional factor $\mu_{c}$ (diagram (b)) for a single leg. The plotted data is for right front (RF) leg, right middle (RM) leg, and right back (RB) leg. The shaded regions are when the robot lifts a certain leg and put it on the next position, and white regions are when the robot pushes its body up. The plot shows failure points (red dashed line), planned curves (blue line), and simulation results (black line). The arrow (green) indicates the margin due to planned safety factor. The results demonstrate a general correspondence of planned and V-rep simulated results.}
		\label{Fig:steps_torque_mu_curve}
\end{figure}

To verify the contact force optimization results, the setup of the robot climbing onto the steps is simulated in V-rep. Fig. \ref{Fig:steps_torque_mu_curve} plots the planned with the simulated torque curves and the critical friction coefficient $\mu_{c}$ as defined by equation (\ref{Eqn:friction_required}) with their failure boundaries for 3 consecutive legs. The torque plotted is the maximal torque among three motors of one leg. When a certain leg is lifted and in the air, the planned torque is set to zero because lifting is achieved by the controller instead of planner. Simulated torques for the lifted leg is negligibly small compared to the torques when it is on the wall so that we can tell lifting phase from these curves. And the frictional factor is also zero with no friction generated for the lifted leg. In Fig. \ref{Fig:steps_torque_mu_curve} $(a)$, the maximal torques of right middle (RM) and right back (RB) legs decrease after right front (RF) leg finished lifting phase (between the shaded interval). With one leg in the air, other legs need to achieve larger contact force to avoid slipping. Once the lifted leg reaches its goal position, contact force would be re-distributed to the 6 legs on the wall. As we can tell, due to the complexity of contact, $\mu_{c}$ tends to exceed the planned values, which is the reason we weighted more on friction than torque. The non-smoothness of the data is in part due to the toe rubbing on the wall (caused by the physics simulator) and to the overshoot of PID body posture controller. Some points of the curves are above the boundaries, but the robot will not slip or over-torque if this is not continuous. 

\subsection{Climbing on the walls while avoiding obstacles}
\begin{figure}[t]
		\centering
		\includegraphics[scale=0.45]{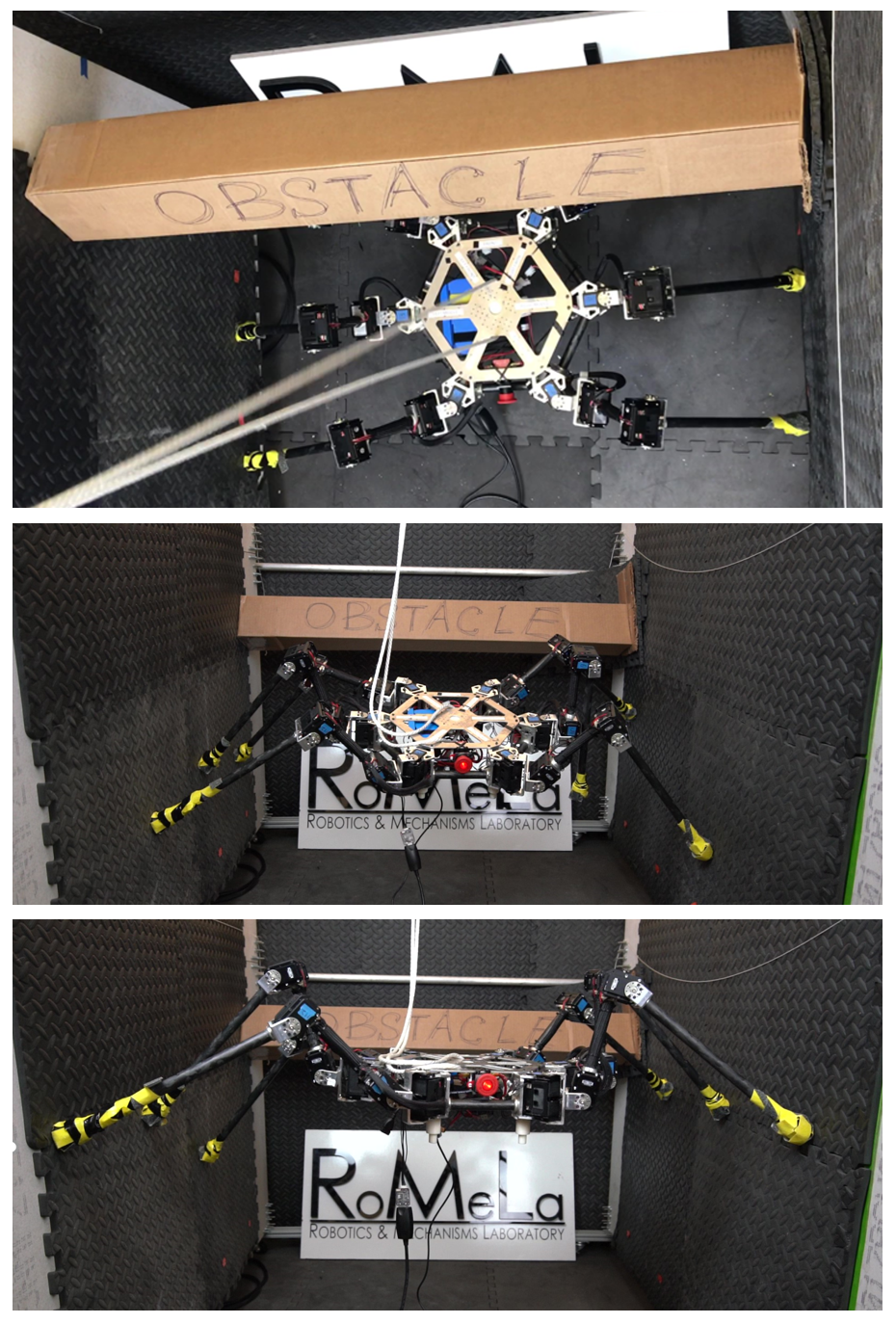}
		\caption{Hardware test for climbing and avoiding an obstacle. The robot starts underneath the obstacle, as shown in the top figure, and then it angles its body and climbs forward and up simultaneously, as shown in the middle and bottom figures.}
		\label{Fig:result_obstacle}
\end{figure}

This scenario focuses on planning the climbing direction and orienting the body to avoid an obstacle between the walls, obstructing a direct path. The planned results are visualized in Fig. \ref{Fig: bracing plot}. The robot doesn't need to rotate its body more than 20 degrees to complete the task; thus, MICP with a linearized body rotation matrix is applied. Due to the obstacle, the feasible contact region shrinks. We manually divide each wall into three convex regions: the upper, middle, and lower rectangle. For each round, the toe position is optimized within one of the three regions selected by the MICP planner. Currently, the robot does not have any vision sensor. In the future, this division can be provided by a perception system, and the complete process will be automated. The hardware test is shown in Fig. \ref{Fig:result_obstacle}.

The MICP plans 8 rounds for this problem. Within each round, the robot lifts each leg once and pushes up the body for 6 times. Hence, the serial convex optimizer plans the force 12 times for each round and 96 times in total. Some statistics for this planner is shown in Table \ref{Tab:MICPsolution_speed} (taken on an Intel Core i7-8750H machine). Since the bulk of the problem is convex, the total solution speed is decently fast. We point out here that a single NLP solver will need to deal with the same amount of variables and constraints; thus, a similar or slower speed is expected (refer to Table I in \cite{winkler2018gait}).    

\begin{table}[]
\caption{Specs for climbing and avoiding obstacle}
\begin{threeparttable}[t]
\begin{center}
\resizebox{1\linewidth}{!}{
\begin{tabular}{@{}clcccc@{}}
\hline
\hline
& Solver & Variables                                                                   & Constraints                                                & T-Solve \tnote{*}                                                      & Total T-Solve          \\ \midrule
\begin{tabular}[c]{@{}c@{}}Posture Planner\\ (MICP - 8 rounds)\end{tabular} & Gurobi & \begin{tabular}[c]{@{}c@{}}480\\ (192 continuous\\ 144 binary)\end{tabular} & 1002                                                         & 420 ms                                                        & \multirow{2}{*}{1380 ms} \\ \cmidrule(r){1-5}
Force Planner                                                        & Gurobi  & \begin{tabular}[c]{@{}c@{}}5856\\ (61 x 96)\end{tabular}                    & \begin{tabular}[c]{@{}c@{}}10368\\ (108 x 96)\end{tabular} & \begin{tabular}[c]{@{}c@{}}960 ms\\ (10 ms x 96)\end{tabular} &                        \\
\hline
\hline
\end{tabular}}
\end{center}
\begin{tablenotes}\footnotesize
\item[*] Include problem set-up time
\end{tablenotes}
\end{threeparttable}
\label{Tab:MICPsolution_speed}
\end{table}

\subsection{Climbing on non-parallel walls}
\begin{figure}[t]
		\centering
		\includegraphics[scale=0.37]{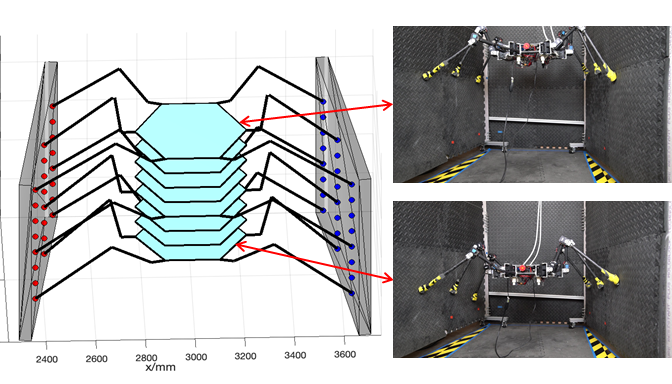}
		\caption {Visualization and hardware test of planning results for climbing on non-parallel walls with $\alpha=20$ degrees.}
		\label{Fig:results_nonparallel}
\end{figure}

\begin{figure}[t]
		\centering
		\makebox[0.45\textwidth][c]{
		\includegraphics[scale=0.45]{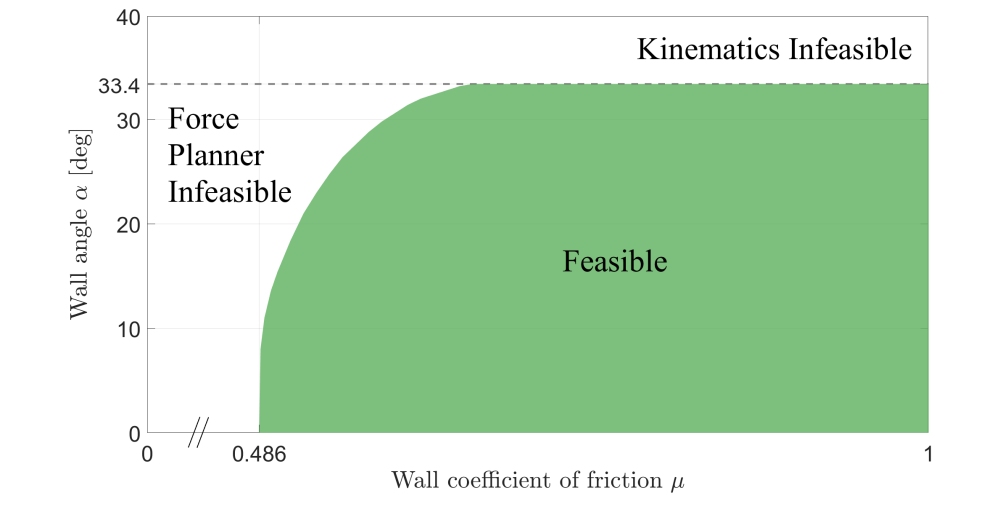}}
		\caption {Feasible region for climbing on non-parallel walls.}
		\label{Fig:feasible_region}
\end{figure}

An interesting variation of the two-wall climbing problem is when the walls are no longer parallel. Ideally, we want the robot to take use of two walls at an arbitrary angle and climb up. This section demonstrates that our planner can be used to plan the climbing motion when two flat walls are at a horizontal angle $\alpha$. We pick $\alpha=20$ degrees and implement the planning results on the hardware, shown in Fig. \ref{Fig:results_nonparallel}. Additionally, we are interested in retrieving a feasible climbing region regarding the given horizontal angle $\alpha$ and the wall coefficient of friction $\mu$. We fix the distance between the two middle legs, and solve this problem by running the planner at discrete grid points for $\alpha$ and $\mu$, and label each point feasible/infeasible. Fig. \ref{Fig:feasible_region} shows the result when the robot is at its own weight without payload (10.3kg). In the plot, the shaded region shows where the robot succeeds, and the rest can be divided into where the robot fails to provide enough force, or fails kinematically (\ie, some toes cannot reach the walls). This result demonstrates that our planner can be extended to broader two-wall cases and possibly to climbing up poles or trees where $\alpha=180$ degrees.


\section{Conclusion and Future Work}
This paper addresses the motion planning problem for a six-legged robot to braces itself between two vertical walls and climbs vertically with end effectors that only use friction. We propose a two-step optimization based planner. The first step is solved as an MICP or NLP. The second part is solved as a series of standard convex optimization problem. We verified the motion plan on three distinct wall profiles.

By decoupling the problem into two parts, we sacrifice some optimality. However, this makes the solver easier to tune. The feasibility region (as shown in \cite{xuan2018multi}) for this problem is pretty narrow. Therefore, it is critical to find the proper weights in the force planner. Since our force planner is convex and interpretable, the difficulty of tuning the weights is attenuated. 

To enable the posture planner to plan according to contact force, we can fuse constraints G through K into the posture planner. This will hopefully enhance the optimality of the solution returned. Additionally, we plan to develop a perception system that maps the wall and retrieves its geometry information. This combined with the planner may automate the complete climbing process and ensure the robot can track the motion plan. Given the solving speed of our algorithm, it could be implemented online to constantly re-plan. We also investigated the problem of climbing spirally up inside a tube with an MICP/NLP solver as the posture planner, which requires the robot to rotate its body by large angles. This will be published in future papers. Other future works include extending the safety factor design principle to other types of grippers \eg, gecko type or microspine, comparing and combining MICP and NLP, \etc. Dynamics could be added into the planner, if dynamic climbing is desired. However, we show that if the planner resolves more than rigid body dynamics, \eg, body compliance, a good option is to decouple the algorithm according to the physics, and resolve them hierarchically.


\section{Acknowledgements}
We would like to thank Julienne Bernal for her contribution to designing and building the climbing walls.

{
\bibliographystyle{ieeetr}
\bibliography{references}
}

\end{document}